\documentclass{article}
\usepackage[margin=0.75in]{geometry}
\usepackage{cite}
\usepackage{amsmath,amssymb,amsfonts}
\usepackage{graphicx}
\usepackage[labelformat=simple]{subfig}
\usepackage{textcomp}
\usepackage{multirow}
\usepackage[linesnumbered,vlined,ruled]{algorithm2e}

\makeatletter
\newcommand\footnoteref[1]{\protected@xdef\@thefnmark{\ref{#1}}\@footnotemark}
\makeatother

\begin{document}

\title{Convolutional Neural Networks for User Identificationbased on Motion Sensors Represented as Image}
\author{Cezara Benegui, {Radu Tudor} Ionescu\\
University of Bucharest, 14 Academiei, Bucharest, Romania}

%\tfootnote{The work of Radu Tudor Ionescu was supported from the EEA Grants 2014-2021, project no. EEA-RO-NO-2018-0496.}

\maketitle

\begin{abstract}
In this paper, we propose a deep learning approach for smartphone user identification based on analyzing motion signals recorded by the accelerometer and the gyroscope, during a single tap gesture performed by the user on the screen. We transform the discrete 3-axis signals from the motion sensors into a gray-scale image representation which is provided as input to a convolutional neural network (CNN) that is pre-trained for multi-class user classification. In the pre-training stage, we benefit from different users and multiple samples per user. After pre-training, we use our CNN as feature extractor, generating an embedding associated to each single tap on the screen. The resulting embeddings are used to train a Support Vector Machines (SVM) model in a few-shot user identification setting, i.e. requiring only 20 taps on the screen during the registration phase. We compare our identification system based on CNN features with two baseline systems, one that employs handcrafted features and another that employs recurrent neural network (RNN) features. All systems are based on the same classifier, namely SVM. To pre-train the CNN and the RNN models for multi-class user classification, we use a different set of users than the set used for few-shot user identification, ensuring a realistic scenario. The empirical results demonstrate that our CNN model yields a top accuracy of 89.75\% in multi-class user classification and a top accuracy of 96.72\% in few-shot user identification. In conclusion, we believe that our system is ready for practical use, having a better generalization capacity than both baselines.
\end{abstract}

\section{Introduction}

Nowadays, common mobile device authentication mechanisms such as PINs, graphical passwords and fingerprint scans offer limited security. These mechanisms are susceptible to guessing (or spoofing in the case of fingerprint scans) and to side channel attacks~\cite{Andriotis-WiSec-2013} such as smudge~\cite{Aviv-WOOT-2010}, reflection~\cite{Xu-CCS-2013,Zhang-SPSM-2012} and video capture attacks~\cite{Simon-SPSM-2013,Shukla-CCS-2014,Ye-NDSS-2017}. On top of this, a fundamental limitation of PINs, passwords, and fingerprint scans is that these mechanisms require explicit user interaction.

Due to the world wide adoption of mobile devices and the advancement of technologies, mobile devices are now equipped with multiple sensors such as accelerometers, gyroscopes, magnetometers, among others. The data recorded by these sensors during the interaction of the user with the mobile device can be used as biometric data to identify the user. Indeed, one-time or continuous user identification based on the data collected by the motion sensors of a mobile device is an actively studied task~\cite{Bo-IPCCC-2014,Buriro-PASSWORDS-2015,Buriro-ISBA-2017,Buriro-SPW-2016,Buriro-CODASPY-2018,Canfora-ICETE-2017,Ehatisham-JNCA-2018,Ku-Access-2019,Li-BIBM-2018,Neverova-Access-2016,Shen-Sensors-2016,Shi-WiMob-2011,Sitova-TIFS-2016,Sun-ECML-2017,Wang-Access-2019}, that emerged after the integration of motion sensors into commonly used mobile devices. 

In this paper, we propose a novel deep learning approach that can identify the user from a single tap on smartphone's touchscreen, using the discrete signals recorded by the accelerometer and the gyroscope during the tap gesture. By minimizing the user's interaction during verification and by removing the requirement to explicitly insert PINs, graphical passwords or scan fingerprints, we eliminate many of the enumerated attacks. 

Our approach is based on transforming the discrete 3-axis signals from the accelerometer and the gyroscope into a gray-scale image representation that can be provided as input for deep convolutional neural networks (CNNs)~\cite{LeCun-IEEE-1998,Krizhevsky-NIPS-2012}. Our image representation is based on repeating the six one-dimensional (1D) signals using a modified version of de Brujin sequences~\cite{Ralston-MM-1982}, such that the $3\times3$ convolutional filters from the first layer of the CNN get to ``see'' every possible tuple of three 1D signals in their receptive field. After transforming the motion signals accordingly, we pre-train several CNN architectures in a multi-class user classification setting. In the pre-training stage, we can leverage the use of data from multiple users and multiple samples per user. After pre-training and selecting the best-performing CNN, we can employ the selected CNN as a deep feature extractor that generates useful embeddings for each tap gesture. The generated embeddings can then be used to train a lightweight model, e.g. Support Vector Machines (SVM)~\cite{Cortes-ML-1995}, to identify the user in a few-shot learning setting. We consider a few-shot learning setting with 20 samples per user in order to enable a fast registration process (20 taps on the screen are enough), similar in terms of time to the registration processes used by standard fingerprint or face authentication systems.

We conduct experiments in order to compare our user identification system based on CNN features with two baselines, one that is based on handcrafted features~\cite{Shen-Sensors-2016} and one that is based on recurrent neural network (RNN) features~\cite{Neverova-Access-2016}. All models are evaluated in a few-shot user identification context using the same classifier, namely SVM. Our CNN model (as well as the baseline RNN model) is pre-trained on a multi-class user classification task. The users involved in the multi-class user classification experiment are different from those involved in the user identification experiment, to simulate a realistic scenario. In order to conduct our experiments, we modify the HMOG data set~\cite{Sitova-TIFS-2016} by extracting shorter signals from the original sessions and by splitting the users in half, using the first half for the preliminary multi-class user classification experiment and the second half for the user identification experiment. Our SVM based on CNN features proves a higher generalization capacity, surpassing both baselines in the user identification experiments. Moreover, according to McNemar's statistical testing~\cite{Dietterich-NC-1998} performed at a confidence level of $0.01$, our improvements over the baselines are significant. With an accuracy of 96.72\%, our SVM based on CNN features seems to be a viable solution for practical usage.

In summary, our contribution is threefold:
\begin{itemize}
\item We propose a novel gray-scale image representation of the discrete signals, designed specifically to be useful as input for CNNs.
\item We propose to pre-train CNNs on a multi-class user classification task in order to obtain useful embeddings for few-shot user identification.
\item We perform comparative experiments showing that our method based on CNN embeddings surpasses both machine learning methods based on handcrafted features and deep learning methods based on RNN embeddings.
\end{itemize}

The rest of this paper is organized as follows. In Section~\ref{sec_Related_Work}, we provide an overview of user identification systems for mobile devices, focusing mainly on systems based on analyzing motion sensors. In Section~\ref{sec_Method}, we present the proposed data representation, our CNN architectures and our user identification model. In Section~\ref{sec_Experiments}, we present the data set, the evaluation metrics and the performed experiments. In Section~\ref{sec_Conclusion}, we conclude our findings and propose some future directions of study.

\section{Related Work}
\label{sec_Related_Work}

The first studies in smartphone user identification used keystroke dynamics~\cite{Clarke-CS-2007,Campisi-SP-2009,Maiorana-SAC-2011}, since the early smartphone devices were equipped with hardware keyboards. To our knowledge, the first study to propose the analysis of accelerometer data in order to recognize the gait of a mobile device user appeared in 2006~\cite{Vildjiounaite-ICPC-2006}. The approach proposed by Vildjiounaite et al.~\cite{Vildjiounaite-ICPC-2006} is to directly measure the similarity between the sample of signal recorded during authentication and a previously-recorded sample of signal that belongs to the user. The samples are compared based on statistical features extracted in the time domain or the frequency domain, without using machine learning. More recent studies explored the task of user identification based on machine learning models~\cite{Bo-IPCCC-2014,Buriro-PASSWORDS-2015,Buriro-ISBA-2017,Buriro-SPW-2016,Buriro-CODASPY-2018,Canfora-ICETE-2017,Ehatisham-JNCA-2018,Ku-Access-2019,Li-BIBM-2018,Neverova-Access-2016,Shen-Sensors-2016,Shi-WiMob-2011,Sitova-TIFS-2016,Sun-ECML-2017,Wang-Access-2019}, attaining better results compared to statistical models such as~\cite{Vildjiounaite-ICPC-2006}. By modeling the user identification task based on motion sensors as a classification task, various models following the standard training and evaluation pipeline used in machine learning can be tested out. The standard machine learning pipeline is essentially based on two steps. The first step is to extract handcrafted features from the discrete motion signals in the time domain or the frequency domain. The second step is to apply a standard machine learning classifier. 

However, some recent works~\cite{Shi-WiMob-2011,Buriro-PASSWORDS-2015,Buriro-SPW-2016,Buriro-ISBA-2017,Sitova-TIFS-2016} have proposed to change the standard pipeline in order to obtain improved performance. For instance, the method proposed by Shi et al.~\cite{Shi-WiMob-2011} uses different modalities for user identification. The approach builds a one-class classifier for each modality and aggregates the results using a meta-classifier. Another approach that employs multiple modalities is proposed by Buriro et al.~\cite{Buriro-PASSWORDS-2015}. Their method is based on motion patterns recorded by the motion sensors and voice patterns recorded by the microphone during a phone call. Some works modify the standard pipeline by adding a feature selection step before classification. The approach described by Sitova et al.~\cite{Sitova-TIFS-2016} uses Principal Component Analysis for feature selection. The authors extracted statistical features specifically designed for tap gestures on the touchscreen of the mobile device. We also analyze tap gestures, but we extract deep CNN features (instead of statistical features) from the motion signals recorded during the taps. The approach proposed by Buriro et al.~\cite{Buriro-SPW-2016} performs user authentication using data recorded by the motion sensors as well as the touchscreen of the mobile device. The authors performed feature selection using the Recursive Feature Elimination method. There are other approaches~\cite{Buriro-ISBA-2017} that use the Information Gain for feature selection.

The methods described in~\cite{Neverova-Access-2016,Sun-ECML-2017} combine the two standard steps (feature extraction and classification) into a single step, by training deep neural networks model in an end-to-end fashion. Similar to these works~\cite{Neverova-Access-2016,Sun-ECML-2017}, we propose an approach based on deep neural networks. Neverova et al.~\cite{Neverova-Access-2016} proposed the use of recurrent connections to model temporal variations of the signals. We take a different approach and propose to use convolutional neural networks in order to obtain deep embeddings of the motion signals recorded by the accelerometer and the gyroscope. It is important to note that Neverova et al.~\cite{Neverova-Access-2016,Sun-ECML-2017} used convolutional neural networks as baselines, showing better results with their recurrent neural networks. While Neverova et al.~\cite{Neverova-Access-2016} used the discrete temporal signals as input for their baseline CNNs, we propose o novel approach to convert the temporal signals into a 2D gray-scale image representation to be used as input for our CNN. As shown in our user identification experiments, the embeddings learned by our CNN are more robust than the embeddings learned by the baseline RNN, leading to significant performance improvements. Another approach based on training neural networks for biometric user identification is proposed by Sun et al.~\cite{Sun-ECML-2017}. While they use recurrent neural networks to recognize users based on keystroke dynamics, we propose to use convolutional neural networks to recognize users based on motion data recorded during screen taps. While works such as~\cite{Neverova-Access-2016,Sun-ECML-2017} require longer interactions from the user, our method is designed to identify the user based on 1.5 seconds of motion signals recorded during a single tap gesture. Furthermore, our method requires only 20 samples (taps) during the user registration phase.

\section{Method}
\label{sec_Method}

\begin{figure}[!th]
\begin{center}
\includegraphics[width=0.38\linewidth]{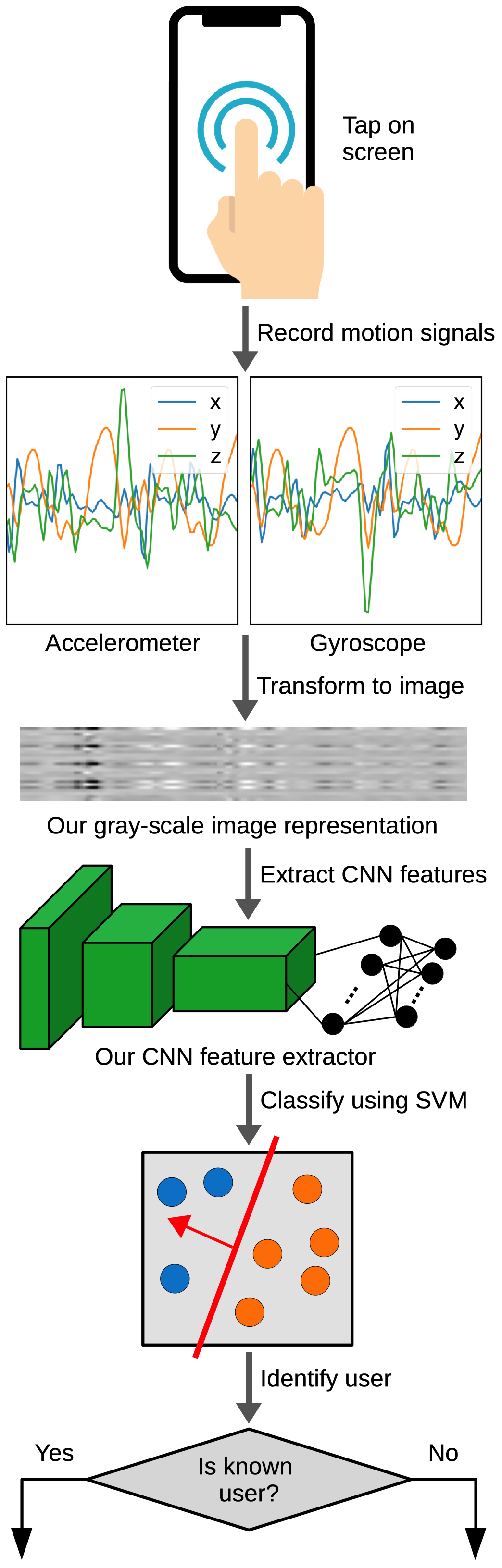}
\end{center}
\vspace{-0.3cm}
\caption{Our user identification pipeline based on analyzing motion signals recorded during the user's screen tap. The signals are combined into a gray-scale image representation which is provided as input to a CNN that is pre-trained on multi-class user classification. An SVM trained on a few examples is used to identify the user. Best viewed in color.}
\vspace{-0.2cm}
\label{fig_pipeline}
\end{figure}

\subsection{Image Representation}

As illustrated in Figure~\ref{fig_pipeline}, our first step is to turn the discrete signals acquired from the smartphone's accelerometer and gyroscope sensors into gray-scale images. We start with six discrete signals of potentially different lengths, represented in the time domain. Although the motion signals are supposed to be recorded at 100 Hz, depending on the processes running on the mobile device, the operating system will not report exactly 100 values per second at perfectly equal time-intervals. Furthermore, the accelerometer and the gyroscope report motion events independently. Although the signals reported for the three axes of a motion sensor are of the same length, the signals reported by two different motion sensors could be of different length. We thus have to normalize the motion signals to a fixed length. Since we record signals for 1.5 seconds at 100 Hz, we expected them to be formed of 150 discrete values. Hence, the signals that are longer or shorter are resized using linear interpolation to a fixed length of 150 values. After resizing a signal, we subtract its minimum magnitude such that each value becomes positive. Each signal is now a vector of 150 positive components. Since the independent signals can have different magnitude scales, e.g. when the motion projected on one axis is much higher in terms of magnitude than the motion projected on another axis, we independently normalize the vectors using the $L_2$-norm. Then, we rescale the values to the interval $[0, 255]$, in order to use the full range of values available for gray-scale images.

\begin{figure}[!th]
\begin{center}
\includegraphics[width=0.4\linewidth]{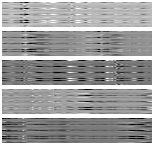}
\end{center}
%\vspace{-0.3cm}
\caption{Gray-scale images of $25 \times 150$ pixels resulted after our conversion of the discrete signals recorded at 100 Hz for 1.5 seconds.}
%\vspace{-0.2cm}
\label{fig_gs_images}
\end{figure}

Next, we choose to consider each 1D discrete signal as a row vector and concatenate the vectors column-wise in specific sequence that allows repetition. The sequence is chosen by taking into consideration that recent CNN architectures, e.g. GoogLeNet~\cite{Szegedy-CVPR-2015} and ResNet~\cite{He-CVPR-2016}, use convolutional filters with a receptive field (spatial support) not higher than $3\times3$. Given that we want to employ such modern design recommendations into our CNN architectures, we need to make sure that every possible tuple of three 1D signals is ``seen'' by the first convolutional layer. We build our sequence of signals based on the principles of de Brujin sequences~\cite{Ralston-MM-1982}. We first associate digits from 0 to 5 to identify our signals in the generated sequence. A \emph{de Brujin sequence} of order n (length of tuples, in our case, $n=3$) on an alphabet $\Sigma$ of size k (number of signals, in our case, $k=6$) is a cyclic sequence in which every possible string of length n (in our case, triplet) on $\Sigma$ (in our case, $\Sigma = \{0,1,2,3,4,5 \}$) occurs exactly once as a contiguous subsequence. Different from de Brujin sequences, we do not need to include triplets of signals that represent rearrangements of triplets that are already included in our sequence. We also do not need triplets of repeating symbols, e.g. $(3,3,3)$. In order to build our sequence of signals we employ a simple Greedy algorithm, which adds the minimum amount of signals to the sequence in order to include a new triplet, which was not previously included in the sequence. While the minimum length of a de Brujin sequence for $n=3$ and $k=6$ is $218$, we obtain a shorter sequence of length $25$. Our sequence is 0, 1, 2, 3, 4, 5, 0, 2, 4, 5, 1, 3, 0, 4, 1, 2, 5, 3, 0, 2, 0, 5, 1, 3, 4. Using the generated sequence, we build our gray-scale image representation of $25 \times 150$ pixels. In Figure~\ref{fig_gs_images}, we illustrate some image representations constructed for a set of randomly selected recording sessions, each of 1.5 seconds in length. After obtaining the image representations, we provide them as the input for the convolutional neural networks described in the following section. 

\subsection{CNN-based Feature Extraction}

Recent methods for object recognition~\cite{Krizhevsky-NIPS-2012,Simonyan-ICLR-14,Szegedy-CVPR-2015,He-CVPR-2016} and other computer vision tasks~\cite{Georgescu-Access-2019,Ren-NIPS-2015,Redmon-CVPR-2016,Ionescu-CVPR-2016,Samala-MP-2016,Wahab-M-2019} are based on deep learning~\cite{LeCun-Nature-2015,Goodfellow-MITPress-2016}. The main approach in this area is represented by CNNs~\cite{Krizhevsky-NIPS-2012,Simonyan-ICLR-14,Szegedy-CVPR-2015}. Convolutional neural networks are a particular type of feed-forward neural networks that are designed to efficiently process images through the use of a special kind of layer inspired by the human visual cortex, namely the \emph{convolutional layer}. Following the success of transfer learning from pre-trained CNNs~\cite{Ionescu-CVPR-2016,Samala-MP-2016,Wahab-M-2019,Yosinski-NIPS-2014}, we consider them as potentially-useful feature extractors for smartphone user identification, given our custom gray-scale image representation derived from motion signals. Instead of considering pre-trained CNN models on ImageNet~\cite{Russakovsky-IJCV-2015} as others~\cite{Ionescu-CVPR-2016,Samala-MP-2016,Wahab-M-2019}, we devise a multi-class user classification task in order to train our models, before transferring them to the user identification task. This ensures that our CNN models are specifically adapted to same kind of input images.

In this work, we propose three CNN architectures of different depths. Each architecture is composed of a different number of convolutional (conv) layers followed by a fixed number of fully-connected (fc) layers. We use Rectified Linear Units (ReLU)~\cite{Nair-ICML-2010} as activation functions on all layers, except for the classification layer which has Softmax activations.

The first CNN architecture is composed of 6 layers, namely 3 conv layers, 2 fc layers and a Softmax classification layer. Each convolutional layer is followed by a max-pooling layer with a pool size of $2\times2$. The first conv layer is composed of 32 filters with a $3\times3$ spatial support. The filters are applied at a stride of 1 and the input is zero-padded to preserve the spatial dimension. The second conv layer is composed of 64 filters, while the third conv layer is composed of 128 filters. As the first conv layer, the filters from the second and the third conv layers have a receptive field of $3\times3$ components and are applied at a stride of 1. The activation maps are zero-padded to preserve the spatial dimension. The third conv layer is followed by 2 fc layers, each of 256 neurons. We use dropout~\cite{Srivastava-JMLR-2014} on each fc layer, with the dropout rate set to 0.4. The last and final layer is a fc layer with 50 neurons, corresponding to the number of classes (users) from our multi-class user classification task. 

The second CNN architecture is composed of 9 layers, namely 6 conv layers, 2 fc layers and a Softmax classification layer. The 9-layer CNN architecture is derived from the 6-layer CNN architecture, by replicating each conv layer exactly once. Therefore, the first and the second conv layers of the 9-layer CNN have 32 filters, as the first conv layer of the 6-layer CNN. Similarly, the third and the fourth conv layers of the 9-layer CNN have 64 filters, as the second conv layer of the 6-layer CNN. The same rule applies to the fifth and the sixth conv layers of the 9-layer CNN, which have 128 filters. In the 9-layer CNN, only the second, the fourth and the sixth conv layers are followed by max-pooling layers with a pool size of $2\times2$. The other layers and parameters are the same as in the 6-layer CNN architecture.

Our third and deepest CNN architecture has 12 layers, namely 9 conv layers, 2 fc layers and a Softmax classification layer. The 12-layer CNN architecture is derived from the 6-layer CNN architecture, by replicating each conv layer exactly twice. The first 3 conv layers contain 32 filters, the following 3 conv layers contain 64 filters and the last 3 conv layers contain 128 filters. In the 12-layer CNN, only the third, the sixth and the ninth conv layers are followed by max-pooling layers. We note that, in the 12-layer CNN architecture, we preserve the other (fc) layers and parameters (stride, kernel size, pool size, dropout rate) from the previously-presented CNN architectures.

All models are trained using the Adam optimizer~\cite{Kingma-ICLR-2015} with the categorical cross-entropy loss function. We train and test our CNN architectures on the multi-class user classification task, in order to select the best-performing architecture. After finding the best CNN model, we remove its Softmax layer and use the activation maps from the last remaining fc layer as feature vectors (deep embeddings). Given that the last fc layer is formed of 256 neurons, we will obtain 256-dimensional embeddings. 

\subsection{Few-shot User Identification}

In a realistic user identification scenario, we do not have access to a set of samples performed by attackers (impersonators) in order to build a binary classification model. Therefore, we can either train an outlier detection model, e.g. a one-class SVM~\cite{Shi-WiMob-2011}, or use a pool of data samples that belong neither to the rightful user nor to the attackers and train a binary classifier. We opt for the second approach and model the user identification task as a binary classification task, in which the positive samples belong to the rightful user and the negative samples belong to a set of users that have nothing in common with the attackers (involved only at test time). In this context, we employ the SVM binary classifier. For binary classification problems, kernel classifiers~\cite{Taylor-CUP-2004}, such as SVM, look for a discriminative function $g$ that assigns positive labels ($+1$) to examples that belong to one class and negative labels ($-1$) to examples that belong to the other class. The function $g$ is linear in the feature space and can be expressed as follows:
\begin{equation}
\begin{split}
g(x)=\mbox{sign}(\langle w, x \rangle + b),
\end{split}
\end{equation}
where $x$ is a feature vector, $w$ and $b$ are the weight vector and the bias term learned by the kernel classifier and $\langle \cdot, \cdot \rangle$ is the dot product. In our case, $x$ is a 256-dimensional feature vector provided a CNN.

Different kernel classifiers may use different criteria to find an optimal vector of weights. The SVM classifier~\cite{Cortes-ML-1995} aims at finding the vector $w$ and the bias $b$ that define the hyperplane which separates the training samples by a maximum margin. Mathematically, the SVM classifier chooses the weights $w$ and the bias term $b$ that satisfy the following optimization criterion:
\begin{equation}
\begin{split}
\min_{w,b}\frac{1}{n}\sum\limits_{i=1}^n[1-y_i(\langle w,x_i \rangle + b)]_+ + C \lVert w \rVert^2 ,
\end{split}
\end{equation}
where $n$ is the number of training samples, $y_i$ is the label ($+1$ or $-1$) of the training example $x_i$, $C$ is a regularization parameter, $[x]_+=\max \lbrace x, 0 \rbrace$ and $\lVert \cdot \rVert^2$ is the $L_2$-norm.

Kernel classifiers rely on a kernel function to embed the data into a high dimensional space, in which non-linear relations become linear. A kernel function captures the intuitive notion of similarity among pairs of data samples from a specific domain, and can be any function defined on the respective domain that is symmetric and positive definite. We opt for two popular kernel functions, namely the linear kernel, which is given by the dot product between pairs of samples, and the Radial Basis Function (RBF) kernel, which is given by the following equation:
\begin{equation}
\begin{split}
k_{RBF}(x_i,x_j)= exp(-\gamma \lVert x_i - x_j \rVert^2 ),
\end{split}
\end{equation}
where $x_i$ and $x_j$ are two data samples, $exp(\cdot)$ is the exponential function and $\gamma$ is a parameter that controls the range of possible output values for the RBF kernel.

\section{Experiments}
\label{sec_Experiments}

\subsection{Data Set}

In order to successfully test the identification of users, we use a data set which contains values from two motion sensors, the accelerometer and the gyroscope, collected from 100 users while sitting~\cite{Sitova-TIFS-2016}. We record the motion sensor values while users perform a single tap on the screen. The recording starts with 0.5 seconds before the tap event and ends with 1 second after the tap event. Both sensors report values on three axes (x, y, z) at about 100 Hz. Hence, an example is composed of six discrete signals, three from each sensor, corresponding to the three axes (x, y, z), respectively. Given that each signal is recorded for 1.5 seconds at about 100 Hz, it is represented by roughly 150 values. For each user, we collect motion signals for the first 200 tap events. In total, we have 20.000 data samples.

The data set is randomly split in two equal parts, such that each half contains a disjoint set of 50 users (users in one half are different from users in the other half). The first part of the data set, containing 50 users (with 200 samples per user), is used to train the neural models for the task of multi-class user classification. We use 160 samples per user for training and the rest of 40 samples per user for validation, corresponding to an $80\%$--$20\%$ split of the data. Hence, there are 8.000 samples for training and 2.000 samples for validation. The first part of the data set is also used for hyperparameter tuning of the neural models and the SVM.

The second part of the data set, containing the other 50 users (with 200 samples per user), is used for the user identification experiments. To simulate a realistic setting, we train a binary classifier for each user, including only 20 positive samples and 100 negative samples. Since we have 50 users, we obtain 50 binary classification problems. The binary models are tested on 100 positive samples and 100 negative samples. When a binary model is trained and tested, we are careful to select the negative training and test examples from disjoint subsets of users. By using a disjoint subset of users during inference, we make sure the classification models do not cheat by making use of features specific to the negative users seen during training. This ensures that our user identification experiments simulate a realistic setting, in which samples coming from potential impersonators are not available during training.

\subsection{Evaluation Metrics}

In order to evaluate the deep learning models on the multi-class user classification task, we employ the classification accuracy. To evaluate our baseline and proposed models on the user identification task, we compute the binary confusion matrix for each user. The binary confusion matrix contains the number of true positives (TP), false positives (FP), false negatives (FN) and true negatives (TN), as shown in Table~\ref{Tab_confusion}. We then extract metrics such as the accuracy (ACC), the false acceptance rate (FAR) and the false rejection rate (FRR).

\begin{table}[!t]
\caption{Confusion matrix (also known as contingency table) of a binary classifier with labels $+1$ or $-1$.\label{Tab_confusion}}
\begin{center}
\begin{tabular}{|l|c|c|c|}
\hline
\multicolumn{2}{|c|}{}	& \multicolumn{2}{c|}{Ground-truth labels} \\
\cline{2-4}
							&	Labels	& $+1$							& $-1$\\
\hline
Classifier				& $+1$	& true positive ($TP$)		& false positive ($FP$) \\
predictions			& $-1$		& false negative ($FN$)	& true negative ($TN$) \\
\hline
\end{tabular}
\end{center}
\end{table}

% The four values are described as follow:
% \begin{itemize}
%   \item True positives represents the total number of correctly predicted positive samples
%   \item False positives represents the total number of incorrect predicted positive samples
%   \item False negatives represents the total number of incorrect predicted negative samples
%   \item True negatives represents the total number of correctly predicted negative samples ~\cite{ DBLP:conf/maics/VisaRRK11}
% \end{itemize}

The \emph{accuracy} of the model is given by the total number of correct predictions divided by the total number of predictions. Thus, the accuracy is given by:
\begin{equation}\label{eq_accuracy}
ACC = \frac{TP + TN}{TP + TN + FP + FN}. 
\end{equation}

The \emph{false acceptance rate} is the ratio between the number of false acceptances and the sum of all negative samples (false positives and true negatives):
\begin{equation}\label{eq_far}
FAR = \frac{FP}{FP + TN}.
\end{equation}

The \emph{false rejection rate} is the ratio between the number of false rejections and the sum of all positives samples (false negatives and true positives):
\begin{equation}\label{eq_frr}
FRR = \frac{FN}{FN + TP}.
\end{equation}

\subsection{Baselines}

In this subsection, we present in detail the baseline methods. The first baseline method is based on a recent work by Shen et al.~\cite{Shen-Sensors-2016}, that uses handcrafted features. The second baseline method is based on another recent work by Neverova et al.~\cite{Neverova-Access-2016}, that uses features learned by Long-Short Term Memory (LSTM) networks~\cite{Hochreiter-NC-1997} with convolutional layers (ConvLSTM). The LSTM is a type of RNN architecture that solves the vanishing gradient problem known to affect RNN models. As for our three CNN models, the ConvLSTM is pre-trained on the multi-class user classification task, then used as feature extractor on the user identification task. In the user identification experiments, both baseline models employ an SVM classifier, for a fair comparison to our CNN model.

\subsubsection{Baseline based on handcrafted features}

The values recorded by the motion sensors cannot be used directly as input for a user classifier, as the signals contain a large amount of noise. In order to extract useful characteristics for a given user, one approach is to compute several statistical features that can embed some particularities of the users from our data set. Therefore, on each of the six discrete 1D signals that form an example, we compute the following handcrafted features, as Shen et al.~\cite{Shen-Sensors-2016}:
\begin{itemize}
  \item \textbf{Mean} \textemdash\space is the mean value of a discrete 1D signal;
  \item \textbf{Min} \textemdash\space is the minimum value of a discrete 1D signal;
  \item	\textbf{Max} \textemdash\space is the maximum value of a discrete 1D signal;
  \item \textbf{Variance} \textemdash\space represents the variance value of a discrete 1D signal;
  \item \textbf{Kurtosis} \textemdash\space describes the width of peak of a discrete 1D signal;
  \item \textbf{Skewness} \textemdash\space represents the orientation of peak of a discrete 1D signal;
  \item \textbf{Quantiles} \textemdash\space are the quantiles of a discrete 1D signal, computed from 30\% to 80\%, increasing by a 10\% step.
\end{itemize}

Moreover, we add to the handcrafted features, Pearson and Kendall's Tau correlation values between all the possible combinations of 1D signal pairs, irrespective of sensor type. An example is thus represented by 72 statistical features (12 features $\times$ 6 signals) and 30 correlation features (2 features $\times$ 15 signal pairs). We provide the handcrafted features as input to an SVM classifier.
  
\subsubsection{Baseline based on ConvLSTM features}

In this section, we present another baseline method, which is based on a pre-trained ConvLSTM as feature extractor. As for our three CNN models, we resize the recorded 1D signals to a fixed length of exactly 150 components. We resize the signals using linear interpolation and obtain an input size of $6 \times 150$ components for the ConvLSTM. Since the LSTM network handles temporal inputs directly, we do not need to convert the temporal signals to gray-scale images, as for the CNN models.

The architecture of the ConvLSTM is composed of 5 layers and is similar in size to our best-performing CNN architecture. The first layer of the model is a convolutional LSTM layer with 32 filters and a kernel size of $1\times3$. In this layer, we use ReLU~\cite{Nair-ICML-2010} as the activation function. The first layer is followed by another convolutional LSTM layer, having the same kernel size and the same activation function, but a higher number of filters, i.e. 64. Typical LSTM architectures have no more than two LSTM layers, which are sufficient to model the temporal variations of the input signals. Therefore, after the second convolutional LSTM layer, the activation maps are flattened. Then, we have a fully-connected layer of 128 neurons with ReLU activations, followed by another fully-connected layer of 256 neurons with ReLU activations. The fifth and final layer is the classification layer, which contains 50 neurons (corresponding to the number of users in the multi-class user classification data set) having Softmax activations. The chosen loss function is the categorical cross-entropy. As for our CNN models, we employ the Adam optimizer~\cite{Kingma-ICLR-2015}.

After training the ConvLSTM model on the multi-class user classification task, we remove the classification layer and use the activation maps from the last fully-connected layer of 256 neurons as features. Hence, we obtain 256 features, which we provide as input to an SVM classifier.

\subsection{Parameter and Implementation Choices}
\label{sec_param_tuning}

For our three CNN models, we set the hyperparameters as described next. We train each CNN model for $50$ epochs using a learning rate of $10^{-3}$. By applying early stopping to prevent overfitting, the training stops after about $40$ epochs for every model. We use dropout~\cite{Srivastava-JMLR-2014} on the two fully-connected layers, using a dropout rate of $0.4$. Regarding the mini-batch size, we experiment with three sizes: $32$, $64$ and $128$. After conducting the experiments on multi-class user classification (presented in Section~\ref{sec_multiclass}), we decided to use the shallower 6-layer CNN with a mini-batch size of $32$ in the subsequent experiments. With the CNN architecture and the mini-batch size fixed, we proceed by trying out thinner or wider architectures which produce $128$ or $512$ features instead of $256$. Here, we opted for the CNN model that produces $256$-dimensional feature vectors. Further details are provided in Section~\ref{sec_multiclass}.

For the baseline ConvLSTM model, we keep the same hyperparameters as for the CNN models, for a fair comparison. We thus set the learning rate of $10^{-3}$ and train the model for $50$ epochs using mini-batches of $32$ samples. By applying early stopping to prevent overfitting, the training stops after $20$ epochs. We use dropout on the two fully-connected layers, using a dropout rate of $0.4$. As the CNN model, the ConvLSTM model produces $256$-dimensional feature vectors.
 
For the SVM model, we try out two kernels, namely the linear kernel and the RBF kernel. For the RBF kernel, we set the parameter $\gamma$ as follows:
\begin{equation}
\gamma = \frac{1}{m \cdot Var(X)},
\end{equation}
where $m$ is the number of features, $X$ is a matrix containing the training data and $Var(\cdot)$ is a function that computes the variance.

In order to obtain optimal results, we adjust the SVM model by tuning the regularization parameter $C$ using grid search on the multi-class user classification data set. As possible values for $C$, we consider values in the set $\{0.1, 1, 10, 100\}$. When using the RBF kernel, the best value for the parameter $C$ is $100$, irrespective of the features (handcrafted or deep). When using the linear kernel, we obtained better results with $C=100$ for the handcrafted features and $C=1$ for the deep (CNN or ConvLSTM) features. The bias term of each SVM model is independently adjusted, such that the difference between the FAR and the FRR is less than 1\%. This ensures a fair comparison in terms of accuracy between models.

While the neural models are implemented in TensorFlow~\cite{Abadi-OSDI-2016}, we employ the Scikit-learn~\cite{Pedregosa-JMLR-2011} implementation of SVM.

\subsection{Multi-class User Classification Results}
\label{sec_multiclass}

\begin{table}[!t]
\caption{Train and validation accuracy rates of our CNN architectures with various depths, on the multi-class user classification task. Each architecture is trained with three different mini-batch sizes. The best results are highlighted in bold.\label{tab_multiclass_batches}}
\begin{center}
\begin{tabular}{|l|c|c|c|}
\hline
\textbf{Model}                          & \textbf{Batch size}   & \multicolumn{2}{|c|}{\textbf{Accuracy}} \\
\cline{3-4}
                                        &                       & \textbf{Training}     & \textbf{Validation} \\ 
\hline
                                        & 32                    & 94.41\%               & \textbf{89.75\%} \\ 
\cline{2-4} 
                                        & 64                    & 91.68\%               & 87.80\% \\
\cline{2-4} 
\multirow{-3}{*}{6-layer CNN}           & 128                   & 90.72\%               & 87.45\% \\
\hline
                                        & 32                    & \textbf{95.88\%}      & 88.45\% \\
\cline{2-4} 
                                        & 64                    & 92.20\%               & 87.95\% \\
\cline{2-4} 
\multirow{-3}{*}{9-layer CNN}           & 128                   & 90.02\%               & 86.00\% \\
\hline
                                        & 32                    & 91.16\%               & 87.20\% \\
\cline{2-4} 
                                        & 64                    & 92.18\%               & 87.85\% \\
\cline{2-4} 
\multirow{-3}{*}{12-layer CNN}          & 128                   & 82.68\%               & 83.95\% \\
\hline
\end{tabular}
\end{center}
\end{table}

\begin{table}[!t]
\caption{Train and validation accuracy rates of our 6-layer CNN architecture with various embedding sizes, on the multi-class user classification task. The best results are highlighted in bold.\label{tab_multiclass_embeddings}}
\begin{center}
\begin{tabular}{|l|c|c|c|}
\hline
\textbf{Model}                          & \textbf{Embedding size}   & \multicolumn{2}{|c|}{\textbf{Accuracy}} \\
\cline{3-4}
                                        &                           & \textbf{Training}     & \textbf{Validation} \\ 
\hline
\multirow{3}{*}{6-layer CNN}            & 128                       & \textbf{97.13\%}      & 89.15\% \\
\cline{2-4} 
                                        & 256                       & 94.41\%               & \textbf{89.75\%} \\
\cline{2-4} 
                                        & 512                       & 93.17\%               & 87.35\% \\
\hline
\end{tabular}
\end{center}
\end{table}

In Table~\ref{tab_multiclass_batches}, we present the results obtained by our three different CNN architectures on the multi-class user classification task. Since neural networks are sensitive to the initialization of the weights, we train each model for 5 times, reporting the average results. We note that, as the architecture grows deeper, the validation accuracy tends to drop slightly. Given that our training data is limited to 8.000 examples, with 160 samples per class, we conclude that the deeper networks are too deep with respect to our training set size. Besides trying different architectures, we also tested various mini-batch sizes: 32, 64 or 128. We hereby note that Jastrzebski et al.~\cite{Jastrzebski-ICANN-2018} observed that the learning rate and the mini-batch size are strongly correlated. Furthermore, they note that similar performance can be obtained with different combinations of learning rate and mini-batch size. Instead of trying various learning rates and mini-batch size combinations, we decided to fix the learning rate to $10^{-3}$ and find the optimal mini-batch size that corresponds to our fixed learning rate. The empirical results presented in Table~\ref{tab_multiclass_batches} indicate that a mini-batch size of 32 is the optimal value from the CNN architectures of 6 and 9 layers, respectively. However, the 12-layer CNN architecture attains better results with mini-batches of 64 samples. While the 9-layer CNN yields a better training accuracy (95.88\%), our 6-layer CNN has a stronger generalization capacity, attaining an accuracy of 89.75\% on the validation set. We thus choose the 6-layer CNN based on mini-batches of 32 samples for the subsequent experiments.

\begin{table}[!t]
\caption{Train and validation accuracy rates of our 6-layer CNN architecture versus the 5-layer ConvLSTM, on the multi-class user classification task. Both models are trained with mini-batches of 32 samples and produce 256-dimensional embeddings. The best results are highlighted in bold.\label{tab_multiclass_lstm}}
\begin{center}
\begin{tabular}{|l|c|c|}
\hline
\textbf{Model}                          & \multicolumn{2}{|c|}{\textbf{Accuracy}} \\
\cline{2-3}
                                        & \textbf{Training}     & \textbf{Validation} \\ 
\hline
5-layer ConvLSTM                        & \textbf{94.93\%}      & 86.70\% \\
\hline
6-layer CNN                             & 94.41\%               & \textbf{89.75\%} \\
\hline
\end{tabular}
\end{center}
\end{table}

\begin{table*}[!t]
\caption{Few-shot user identification results provided by our SVM based on CNN features versus two baselines, an SVM based on handcrafted features and an SVM based on ConvLSTM features. Results are reported for two kernel functions, linear and RBF, and the regularization parameter C which is determined using grid search on the multi-class user classification experiments. The reported accuracy, FAR and FRR values represent the average values determined on the 50 users involved in the few-shot user identification task. The marker $*$ indicates that the performance is significantly better than the SVM baseline based on handcrafted features, according to a paired McNemar's test performed at a significance level of 0.01. The marker $\diamond$ indicates that the performance is significantly better than the SVM baseline based on ConvLSTM features, according to a paired McNemar's test performed at a significance level of 0.01.\label{tab_user_identification}}
\begin{center}
\begin{tabular}{|l|c|c|c|c|c|}
\hline
\multicolumn{1}{|l|}{\textbf{Model}}                    & \textbf{Kernel}   & \textbf{C} & \textbf{Accuracy} & \textbf{FAR} & \textbf{FRR} \\
\hline
\multirow{2}{*}{Handcrafted features + SVM} & linear        & 100           & 85.60\%  & 14.66\%  & 14.12\%  \\
\cline{2-6} 
                                            & RBF           & 100           & 87.94\%     &  12.16\%       & 11.93\%   \\
\hline
\multirow{2}{*}{ConvLSTM features + SVM}    & linear        & 1             & 94.68\%$^{*}$      &   5.34\%$^{*}$     & 5.29\%$^{*}$  \\
\cline{2-6} 
                                            & RBF           & 100           & 94.47\%$^{*}$      &    5.08\%$^{*}$     & 5.98\%$^{*}$  \\
\hline
\multirow{2}{*}{CNN features + SVM}         & linear        & 1             & \textbf{96.72\%}$^{*,\diamond}$ & \textbf{3.10\%}$^{*,\diamond}$ & \textbf{3.45\%}$^{*,\diamond}$    \\       
\cline{2-6} 
                                            & RBF           & 100           & 96.28\%$^{*,\diamond}$      & 3.62\%$^{*,\diamond}$       & 3.81\%$^{*,\diamond}$    \\
\hline
\end{tabular}
\end{center}
\end{table*}

After setting out the CNN architecture, we conducted further experiments to determine the optimal size of the embedding. In addition to the original 6-layer CNN that produces 256-dimensional embeddings, we try out a thinner model producing 128-dimensional embeddings and a wider model producing 512-dimensional embeddings. The corresponding results are presented in Table~\ref{tab_multiclass_embeddings}. We note that Table~\ref{tab_multiclass_embeddings} contains the average accuracy rates computed over 5 runs for each model. While the thinner CNN seems to fit better on the training set, yielding an accuracy of 97.13\%, it does not surpass the CNN producing 256-dimensional embeddings, on the validation set. The wider CNN attains the lowest accuracy rates on both training and validation sets. In conclusion, we decided to stick with the CNN architecture that gives us 256-dimensional feature vectors.

Our final aim is to use our best CNN model as a pre-trained feature extractor for the user identification task. For a fair comparison, we apply the ConvLSTM in a similar way, i.e. as a pre-trained feature extractor. Therefore, the first step is to train the ConvLSTM on the multi-class user classification task. The corresponding accuracy rates are presented in Table~\ref{tab_multiclass_lstm}. It is important to note, once again, that the hyperparameters of the ConvLSTM are similar to our best 6-layer CNN (see Section~\ref{sec_param_tuning}). Compared to our CNN, it seems that the LSTM units are not able to properly capture the particularities of the discrete temporal signals. The validation accuracy of the ConvLSTM (86.70\%) is 3.05\% below the validation accuracy of our CNN (89.75\%). As it currently seems, our CNN is a model with higher generalization capacity than the ConvLSTM. Indeed, the difference between the training accuracy and the validation accuracy is lower for our model. %However, as we are about to see in the user identification experiments presented in Section~\ref{sec_user_id}, the ConvLSTM overfits to the set of 50 users used in the multi-class user classification task, attaining lower performance in user identification compared to our CNN model.

\subsection{User Identification Results}
\label{sec_user_id}

In Table~\ref{tab_user_identification}, we present the comparative results on the few-shot user identification task. Our SVM classifier based on CNN features is compared with two baseline SVM classifiers, one based on handcrafted features and one based on ConvLSTM features. For each SVM model, we experiment with two kernel functions, the linear kernel and the RBF kernel. 

With respect to the kernel functions, we note that the RBF kernel gives better results in combination with the handcrafted features, while the linear kernel gives better results in combination with the CNN and the ConvLSTM features. Since the number of handcrafted features (72) is smaller than the number of training samples (120), the classification problem is likely not linearly separable (because we have less features than data samples). Hence, the SVM based on handcrafted features benefits from the use of the RBF kernel, which is known to embed the features in a higher-dimensional space, in which samples can be linearly separated. As the feature vectors provided by the CNN and the ConvLSTM contain 256 features, the 120 training samples are already linearly separable (because we have more features than data samples). Further increasing the feature space through the use of the RBF kernel, seems to be a typical case of the Hughes phenomenon, i.e. the models start to suffer from the curse of dimensionality~\cite{Trunk-PAMI-1979}.

With respect to the features, we note that the SVM based on handcrafted features attains accuracy rates between 85\% and 88\%. We believe that these accuracy rates are not high enough for the system to be used in practice. The SVM based on ConvLSTM features yields accuracy rates between 94\% and 95\%, while our SVM model based on CNN features surpasses both baselines, attaining accuracy rates between 96\% and 97\%. We believe that the performance gap between the ConvLSTM features and our CNN features is caused by the fact that the ConvLSTM model tends to overfit to the training set. Interestingly, our results confirm the recent trends from the deep learning community, advocating in favor of using alternative approaches instead of RNN and LSTM architectures in order to model temporal data, e.g. by employing attention mechanisms~\cite{Vaswani-NIPS-2017}.

%We are particularly surprised to find out that the ConvLSTM features have such a low generalization capacity, despite of attaining top performance in the multi-class user classification experiments. 
%We believe that the performance gap of the ConvLSTM features is caused by overfitting to the set of users used in the ConvLSTM training stage. To test this hypothesis, we conducted few-shot user identification experiments on the same set of users, using only the ConvLSTM model. Since the accuracy of the ConvLSTM in this scenario is 88.82\%, our hypothesis is confirmed. We stress out, however, that it is unrealistic to train and evaluate a model on the same set of users, since in practice, we may not have access to sufficient data samples from each user. We thus conclude that the SVM based on ConvLSTM features is not suitable for practical user identification. Our SVM model based on CNN features surpasses both baselines, attaining accuracy rates between 96\% and 97\%. 

We also performed McNemar's statistical tests~\cite{Dietterich-NC-1998}, at a confidence level of $0.01$, in order to determine if the differences between the three types of features (handcrafted, ConvLSTM and CNN) are statistically significant. Without exception, the accuracy rates reached by our SVM model based on CNN features are significantly better than both baselines. We also note that the baseline based on ConvLSTM features is significantly better than the baseline based on handcrafted features.

\section{Conclusions}
\label{sec_Conclusion}

In this paper, we have presented an approach based on pre-trained CNN features that can identify (authenticate) users by analyzing data recorded by motion sensors incorporated in mobile devices, while the user performs a single tap gesture on the screen. Our approach is based on transforming the discrete signals from motion sensors into a gray-scale image representation which is then provided as input to a convolutional neural network (CNN) that is pre-trained on a multi-class user classification task. After pre-training, we used the CNN as feature extractor, generating an embedding associated to each single tap on the screen. We compared our user identification system based on CNN features with two baseline systems, one that employs handcrafted features and another that employs ConvLSTM features. All systems are based on the SVM classifier, for a fair comparison. To pre-train the CNN and the ConvLSTM models for multi-class user classification, we used a different set of users than the set used for few-shot user identification, ensuring a realistic scenario. The empirical results demonstrate that $(i)$ our system attains a top accuracy of 96.72\% with a FAR of 3.10\% and a FRR of 3.45\%, using only 20 samples per user during training, and $(ii)$ our system is significantly better compared to the considered baselines. We thus conclude that our SVM model based on pre-trained CNN features is suitable for practical usage, having a high accuracy rate while requiring only 20 taps from the user during registration.

In future work, we aim to combine the compared models into an ensemble model that should be able to further improve the identification performance of users based on motion patterns. Here, we could explore various ensemble learning approaches. We also aim to add an attention mechanism to our CNN model, which could further improve its performance. We also consider creating and improving authentication systems by implementing our system as a passive factor in a two-factor authentication scheme.

\bibliographystyle{IEEEtran}
\bibliography{IEEEabrv,references}

\begin{thebibliography}{10}
\providecommand{\url}[1]{#1}
\csname url@samestyle\endcsname
\providecommand{\newblock}{\relax}
\providecommand{\bibinfo}[2]{#2}
\providecommand{\BIBentrySTDinterwordspacing}{\spaceskip=0pt\relax}
\providecommand{\BIBentryALTinterwordstretchfactor}{4}
\providecommand{\BIBentryALTinterwordspacing}{\spaceskip=\fontdimen2\font plus
\BIBentryALTinterwordstretchfactor\fontdimen3\font minus
  \fontdimen4\font\relax}
\providecommand{\BIBforeignlanguage}[2]{{%
\expandafter\ifx\csname l@#1\endcsname\relax
\typeout{** WARNING: IEEEtran.bst: No hyphenation pattern has been}%
\typeout{** loaded for the language `#1'. Using the pattern for}%
\typeout{** the default language instead.}%
\else
\language=\csname l@#1\endcsname
\fi
#2}}
\providecommand{\BIBdecl}{\relax}
\BIBdecl

\bibitem{Andriotis-WiSec-2013}
P.~Andriotis, T.~Tryfonas, G.~Oikonomou, and C.~Yildiz, ``{A Pilot Study on the
  Security of Pattern Screen-Lock Methods and Soft Side Channel Attacks},'' in
  \emph{Proceedings of WiSec}, 2013, pp. 1--6.

\bibitem{Aviv-WOOT-2010}
A.~J. Aviv, K.~Gibson, E.~Mossop, M.~Blaze, and J.~M. Smith, ``{Smudge Attacks
  on Smartphone Touch Screens},'' in \emph{Proceedings of WOOT}, 2010, pp.
  1--7.

\bibitem{Xu-CCS-2013}
Y.~Xu, J.~Heinly, A.~M. White, F.~Monrose, and J.-M. Frahm, ``{Seeing double:
  Reconstructing obscured typed input from repeated compromising
  reflections},'' in \emph{Proceedings of CCS}, 2013, pp. 1063--1074.

\bibitem{Zhang-SPSM-2012}
Y.~Zhang, P.~Xia, J.~Luo, Z.~Ling, B.~Liu, and X.~Fu, ``Fingerprint attack
  against touch-enabled devices,'' in \emph{Proceedings of SPSM}, 2012, pp.
  57--68.

\bibitem{Simon-SPSM-2013}
L.~Simon and R.~Anderson, ``{Pin skimmer: Inferring PINs through the camera and
  microphone},'' in \emph{Proceedings of SPSM}, 2013, pp. 67--78.

\bibitem{Shukla-CCS-2014}
D.~Shukla, R.~Kumar, A.~Serwadda, and V.~V. Phoha, ``Beware, your hands reveal
  your secrets!'' in \emph{Proceedings of CCS}, 2014, pp. 904--917.

\bibitem{Ye-NDSS-2017}
G.~Ye, Z.~Tang, D.~Fang, X.~Chen, K.~I. Kim, B.~Taylor, and Z.~Wang,
  ``{Cracking Android Pattern Lock in Five Attempts},'' in \emph{Proceedings of
  NDSS}, 2017.

\bibitem{Bo-IPCCC-2014}
C.~Bo, L.~Zhang, T.~Jung, J.~Han, X.-Y. Li, and Y.~Wang, ``{Continuous User
  Identification via Touch and Movement Behavioral Biometrics},'' in
  \emph{Proceedings of IPCCC}, 2014, pp. 1--8.

\bibitem{Buriro-PASSWORDS-2015}
A.~Buriro, B.~Crispo, F.~Del~Frari, J.~Klardie, and K.~Wrona, ``{ITSME:
  Multi-modal and Unobtrusive Behavioural User Authentication for
  Smartphones},'' in \emph{Proceedings of PASSWORDS}, 2015, pp. 45--61.

\bibitem{Buriro-ISBA-2017}
A.~Buriro, B.~Crispo, and Y.~Zhauniarovich, ``{Please Hold On: Unobtrusive User
  Authentication using Smartphone's built-in Sensors},'' in \emph{Proceedings
  of ISBA}, 2017, pp. 1--8.

\bibitem{Buriro-SPW-2016}
A.~Buriro, B.~Crispo, F.~Delfrari, and K.~Wrona, ``Hold and sign: A novel
  behavioral biometrics for smartphone user authentication,'' in
  \emph{Proceedings of SPW}, 2016, pp. 276--285.

\bibitem{Buriro-CODASPY-2018}
A.~Buriro, B.~Crispo, S.~Gupta, and F.~Del~Frari, ``{DIALERAUTH: A
  Motion-assisted Touch-based Smartphone User Authentication Scheme},'' in
  \emph{Proceedings of CODASPY}, 2018, pp. 267--276.

\bibitem{Canfora-ICETE-2017}
G.~Canfora, P.~di~Notte, F.~Mercaldo, and C.~A. Visaggio, ``{A Methodology for
  Silent and Continuous Authentication in Mobile Environment},'' in
  \emph{Proceedings of ICETE}, M.~S. Obaidat, Ed., 2017, pp. 241--265.

\bibitem{Ehatisham-JNCA-2018}
M.~Ehatisham-ul Haq, M.~A. Azam, U.~Naeem, Y.~Amin, and J.~Loo, ``Continuous
  authentication of smartphone users based on activity pattern recognition
  using passive mobile sensing,'' \emph{Journal of Network and Computer
  Applications}, vol. 109, pp. 24--35, 2018.

\bibitem{Ku-Access-2019}
Y.~Ku, L.~H. Park, S.~Shin, and T.~Kwon, ``Draw it as shown: Behavioral pattern
  lock for mobile user authentication,'' \emph{IEEE Access}, vol.~7, pp.
  69\,363--69\,378, 2019.

\bibitem{Li-BIBM-2018}
H.~Li, J.~Yu, and Q.~Cao, ``{Intelligent Walk Authentication: Implicit
  Authentication When You Walk with Smartphone},'' in \emph{Proceedings of
  BIBM}, 2018, pp. 1113--1116.

\bibitem{Neverova-Access-2016}
N.~Neverova, C.~Wolf, G.~Lacey, L.~Fridman, D.~Chandra, B.~Barbello, and
  G.~Taylor, ``{Learning Human Identity from Motion Patterns},'' \emph{IEEE
  Access}, vol.~4, pp. 1810--1820, 2016.

\bibitem{Shen-Sensors-2016}
C.~Shen, T.~Yu, S.~Yuan, Y.~Li, and X.~Guan, ``{Performance Analysis of
  Motion-Sensor Behavior for User Authentication on Smartphones},''
  \emph{Sensors}, vol.~16, no.~3, p. 345, 2016.

\bibitem{Shi-WiMob-2011}
W.~Shi, J.~Yang, Y.~Jiang, F.~Yang, and Y.~Xiong, ``{SenGuard: Passive User
  Identification on Smartphones Using Multiple Sensors},'' in \emph{Proceedings
  of WiMob}, 2011, pp. 141--148.

\bibitem{Sitova-TIFS-2016}
Z.~Sitov{\'a}, J.~{\v{S}}edenka, Q.~Yang, G.~Peng, G.~Zhou, P.~Gasti, and K.~S.
  Balagani, ``{HMOG: New Behavioral Biometric Features for Continuous
  Authentication of Smartphone Users},'' \emph{IEEE Transactions on Information
  Forensics and Security}, vol.~11, no.~5, pp. 877--892, 2016.

\bibitem{Sun-ECML-2017}
L.~Sun, Y.~Wang, B.~Cao, S.~Y. Philip, W.~Srisa-An, and A.~D. Leow,
  ``Sequential keystroke behavioral biometrics for mobile user identification
  via multi-view deep learning,'' in \emph{Proceedings of ECML-PKDD}, 2017, pp.
  228--240.

\bibitem{Wang-Access-2019}
R.~Wang and D.~Tao, ``{Context-Aware Implicit Authentication of Smartphone
  Users Based on Multi-Sensor Behavior},'' \emph{IEEE Access}, vol.~7, pp.
  119\,654--119\,667, 2019.

\bibitem{LeCun-IEEE-1998}
Y.~LeCun, L.~Bottou, Y.~Bengio, and P.~Haffner, ``{Gradient-based Learning
  Applied to Document Recognition},'' \emph{Proceedings of the IEEE}, vol.~86,
  no.~11, pp. 2278--2324, 1998.

\bibitem{Krizhevsky-NIPS-2012}
A.~Krizhevsky, I.~Sutskever, and G.~E. Hinton, ``{ImageNet Classification with
  Deep Convolutional Neural Networks},'' in \emph{Proceedings of NIPS}, 2012,
  pp. 1097--1105.

\bibitem{Ralston-MM-1982}
A.~Ralston, ``{De Bruijn Sequences—-A Model Example of the Interaction of
  Discrete Mathematics and Computer Science},'' \emph{Mathematics Magazine},
  vol.~55, no.~3, pp. 131--143, 1982.

\bibitem{Cortes-ML-1995}
C.~Cortes and V.~Vapnik, ``{Support-Vector Networks},'' \emph{Machine
  Learning}, vol.~20, no.~3, pp. 273--297, 1995.

\bibitem{Dietterich-NC-1998}
T.~G. Dietterich, ``{Approximate Statistical Tests for Comparing Supervised
  Classification Learning Algorithms},'' \emph{Neural Computation}, vol.~10,
  no.~7, pp. 1895--1923, 1998.

\bibitem{Clarke-CS-2007}
N.~L. Clarke and S.~Furnell, ``Advanced user authentication for mobile
  devices,'' \emph{Computers \& Security}, vol.~26, no.~2, pp. 109--119, 2007.

\bibitem{Campisi-SP-2009}
P.~Campisi, E.~Maiorana, M.~Lo~Bosco, and A.~Neri, ``User authentication using
  keystroke dynamics for cellular phones,'' \emph{IET Signal Processing},
  vol.~3, no.~4, pp. 333--341, 2009.

\bibitem{Maiorana-SAC-2011}
E.~Maiorana, P.~Campisi, N.~Gonz{\'a}lez-Carballo, and A.~Neri, ``Keystroke
  dynamics authentication for mobile phones,'' in \emph{Proceedings of SAC},
  2011, pp. 21--26.

\bibitem{Vildjiounaite-ICPC-2006}
E.~Vildjiounaite, S.-M. M{\"a}kel{\"a}, M.~Lindholm, R.~Riihim{\"a}ki,
  V.~Kyll{\"o}nen, J.~M{\"a}ntyj{\"a}rvi, and H.~Ailisto, ``Unobtrusive
  multimodal biometrics for ensuring privacy and information security with
  personal devices,'' in \emph{Proceedings of PERVASIVE}, 2006, pp. 187--201.

\bibitem{Szegedy-CVPR-2015}
C.~Szegedy, W.~Liu, Y.~Jia, P.~Sermanet, S.~Reed, D.~Anguelov, D.~Erhan,
  V.~Vanhoucke, and A.~Rabinovich, ``{Going Deeper With Convolutions},'' in
  \emph{Proceedings of CVPR}, 2015, pp. 1--9.

\bibitem{He-CVPR-2016}
K.~He, X.~Zhang, S.~Ren, and J.~Sun, ``{Deep Residual Learning for Image
  Recognition},'' in \emph{Proceedings of CVPR}, 2016, pp. 770--778.

\bibitem{Simonyan-ICLR-14}
K.~Simonyan and A.~Zisserman, ``{Very Deep Convolutional Networks for
  Large-Scale Image Recognition},'' in \emph{Proceedings of ICLR}, 2014.

\bibitem{Georgescu-Access-2019}
M.-I. Georgescu, R.~T. Ionescu, and M.~Popescu, ``{Local Learning with Deep and
  Handcrafted Features for Facial Expression Recognition},'' \emph{IEEE
  Access}, vol.~7, pp. 64\,827--64\,836, 2019.

\bibitem{Ren-NIPS-2015}
S.~Ren, K.~He, R.~Girshick, and J.~Sun, ``{Faster R-CNN: Towards Real-Time
  Object Detection with Region Proposal Networks},'' in \emph{Proceedings of
  NIPS}, 2015, pp. 91--99.

\bibitem{Redmon-CVPR-2016}
J.~Redmon, S.~Divvala, R.~Girshick, and A.~Farhadi, ``{You Only Look Once:
  Unified, Real-Time Object Detection},'' in \emph{Proceedings of CVPR}, 2016,
  pp. 779--788.

\bibitem{Ionescu-CVPR-2016}
R.~T. Ionescu, B.~Alexe, M.~Leordeanu, M.~Popescu, D.~Papadopoulos, and
  V.~Ferrari, ``{How hard can it be? Estimating the difficulty of visual search
  in an image},'' in \emph{Proceedings of CVPR}, 2016, pp. 2157--2166.

\bibitem{Samala-MP-2016}
R.~K. Samala, H.-P. Chan, L.~Hadjiiski, M.~A. Helvie, J.~Wei, and K.~Cha,
  ``Mass detection in digital breast tomosynthesis: Deep convolutional neural
  network with transfer learning from mammography,'' \emph{Medical Physics},
  vol.~43, no.~12, pp. 6654--6666, 2016.

\bibitem{Wahab-M-2019}
N.~Wahab, A.~Khan, and Y.~S. Lee, ``{Transfer learning based deep CNN for
  segmentation and detection of mitoses in breast cancer histopathological
  images},'' \emph{Microscopy}, vol.~68, no.~3, pp. 216--233, 2019.

\bibitem{LeCun-Nature-2015}
Y.~LeCun, Y.~Bengio, and G.~Hinton, ``Deep learning,'' \emph{Nature}, vol. 521,
  no. 7553, pp. 436--444, 05 2015.

\bibitem{Goodfellow-MITPress-2016}
\BIBentryALTinterwordspacing
I.~Goodfellow, A.~Courville, and Y.~Bengio, \emph{Deep Learning}.\hskip 1em
  plus 0.5em minus 0.4em\relax MIT Press, 2016. [Online]. Available:
  \url{http://www.deeplearningbook.org}
\BIBentrySTDinterwordspacing

\bibitem{Yosinski-NIPS-2014}
J.~Yosinski, J.~Clune, Y.~Bengio, and H.~Lipson, ``How transferable are
  features in deep neural networks?'' in \emph{Proceedings of NIPS}, 2014, pp.
  3320--3328.

\bibitem{Russakovsky-IJCV-2015}
O.~Russakovsky, J.~Deng, H.~Su, J.~Krause, S.~Satheesh, S.~Ma, Z.~Huang, K.~A.,
  A.~Khosla, M.~Bernstein, A.~C. Berg, and L.~Fei-Fei, ``{ImageNet Large Scale
  Visual Recognition Challenge},'' \emph{International Journal of Computer
  Vision}, vol. 115, no.~3, pp. 211--252, 2015.

\bibitem{Nair-ICML-2010}
V.~Nair and G.~E. Hinton, ``{Rectified Linear Units Improve Restricted
  Boltzmann Machines},'' in \emph{Proceedings of ICML}, 2010, pp. 807--814.

\bibitem{Srivastava-JMLR-2014}
N.~Srivastava, G.~Hinton, A.~Krizhevsky, I.~Sutskever, and R.~Salakhutdinov,
  ``{Dropout: A Simple Way to Prevent Neural Networks from Overfitting},''
  \emph{Journal of Machine Learning Research}, vol.~15, pp. 1929--1958, 2014.

\bibitem{Kingma-ICLR-2015}
D.~P. Kingma and J.~Ba, ``Adam: A method for stochastic optimization,'' in
  \emph{Proceedings of ICLR}, 2015.

\bibitem{Taylor-CUP-2004}
J.~Shawe-Taylor and N.~Cristianini, \emph{{Kernel Methods for Pattern
  Analysis}}.\hskip 1em plus 0.5em minus 0.4em\relax Cambridge University
  Press, 2004.

\bibitem{Hochreiter-NC-1997}
S.~Hochreiter and J.~Schmidhuber, ``{Long Short-term Memory},'' \emph{Neural
  Computation}, vol.~9, pp. 1735--80, 1997.

\bibitem{Abadi-OSDI-2016}
M.~Abadi, P.~Barham, J.~Chen, Z.~Chen, A.~Davis, J.~Dean, M.~Devin,
  S.~Ghemawat, G.~Irving, M.~Isard, M.~Kudlur, J.~Levenberg, R.~Monga,
  S.~Moore, D.~G. Murray, B.~Steiner, P.~Tucker, V.~Vasudevan, P.~Warden,
  M.~Wicke, Y.~Yu, and X.~Zheng, ``{TensorFlow: A system for large-scale
  machine learning},'' in \emph{Proceedings of OSDI}, 2016, pp. 265--283.

\bibitem{Pedregosa-JMLR-2011}
F.~Pedregosa, G.~Varoquaux, A.~Gramfort, V.~Michel, B.~Thirion, O.~Grisel,
  M.~Blondel, P.~Prettenhofer, R.~Weiss, V.~Dubourg \emph{et~al.},
  ``{Scikit-learn: Machine learning in Python},'' \emph{Journal of Machine
  Learning Research}, vol.~12, pp. 2825--2830, 2011.

\bibitem{Jastrzebski-ICANN-2018}
S.~Jastrzebski, Z.~Kenton, D.~Arpit, N.~Ballas, A.~Fischer, Y.~Bengio, and
  A.~Storkey, ``{Width of Minima Reached by Stochastic Gradient Descent is
  Influenced by Learning Rate to Batch Size Ratio},'' in \emph{Proceedings of
  ICANN}, vol. 11141, 2018, pp. 392--402.

\bibitem{Trunk-PAMI-1979}
G.~V. Trunk, ``A problem of dimensionality: A simple example,'' \emph{IEEE
  Transactions on Pattern Analysis and Machine Intelligence}, no.~3, pp.
  306--307, 1979.

\bibitem{Vaswani-NIPS-2017}
A.~Vaswani, N.~Shazeer, N.~Parmar, J.~Uszkoreit, L.~Jones, A.~N. Gomez,
  {\L}.~Kaiser, and I.~Polosukhin, ``Attention is all you need,'' in
  \emph{Proceedings of NIPS}, 2017, pp. 5998--6008.

\end{thebibliography}

\end{document}